# ISTD-YOLO: A Multi-Scale Lightweight High-Performance Infrared Small Target Detection Algorithm

Shang Zhang[1,2,3]*, Yujie Cui[1,2,3], Ruoyan Xiong[1,2,3], and Huanbin Zhang[1,2,3]

[1] College of Computer and Information Technology, China Three Gorges University, Hubei, Yichang, 443002, China
[2] Hubei Province Engineering Technology Research Center for Construction Quality Testing Equipment, China Three Gorges University, Yichang 443002, China
[3] Hubei Key Laboratory of Intelligent Vision Based Monitoring for Hydroelectric Engineering, China Three Gorges University, Yichang 443002, China
`zhangshang@ctgu.edu.cn`

**Abstract.** Aiming at the detection difficulties of infrared images such as complex background, low signal-to-noise ratio, small target size and weak brightness, a lightweight infrared small target detection algorithm ISTD-YOLO based on improved YOLOv7 was proposed. Firstly, the YOLOv7 network structure was lightweight reconstructed, and a three-scale lightweight network architecture was designed. Then, the ELAN-W module of the model neck network is replaced by VoV-GSCSP to reduce the computational cost and the complexity of the network structure. Secondly, a parameter-free attention mechanism was introduced into the neck network to enhance the relevance of local context information. Finally, the Normalized Wasserstein Distance (NWD) was used to optimize the commonly used IoU index to enhance the localization and detection accuracy of small targets. Experimental results show that compared with YOLOv7 and the current mainstream algorithms, ISTD-YOLO can effectively improve the detection effect, and all indicators are effectively improved, which can achieve high-quality detection of infrared small targets.

## 1 Introduction

Infrared small target detection is an important research direction in the field of computer vision. It is a method to use infrared thermal imaging technology to detect weak and difficult to identify targets in infrared images. Due to the weak brightness and small size of infrared dim target itself, it is difficult to detect it directly because of the lack of obvious shape, texture and color information [1]. Secondly, in the real scene, false alarm is also a difficult problem to solve. Therefore, accurate and fast recognition of dim small targets from infrared images is the next focus of research.

Traditional object detection technologies such as Viola Jones Detectors, Histogram of Oriented Gradients, Deformable Parts Model, etc., can detect some simple objects effectively. However, the detection effect of complex objects is relatively poor. Moreover, they mainly rely on manually designed feature extraction and classification algorithms, and need to be specifically adjusted and optimized for different goals, which has great limitations.

*Corresponding author.

Nowadays, deep learning is developing rapidly, and target detection technology is becoming more mature. For the above problems of infrared small target detection, it can be solved by deep learning. Object detection algorithms based on deep learning can be roughly divided into the following four categories:

1) Two-stage object detection algorithm, which is a candidate region-based algorithm represented by R-CNN series [2, 3, 4], with detection ranging from coarse to fine.

2) Single-stage object detection algorithms, represented by YOLO series [5, 6, 7] and SSD[8], are detection and recognition algorithms based on the regression framework. These algorithms directly use the entire image as input to a neural network to predict the class of the object and the position of the bounding box.

3) Frameless target detection algorithm is based on key point and center area represented by CenterNet[9, 10]. This algorithm abandons the complex and inefficient anchor frame operation, so it is relatively simple.

4) Target detection algorithm based on attention mechanism is a class of algorithms represented by transformer. The main idea is to treat target detection as a sequence-to-sequence (Seq2Seq) problem and use transformer to solve it.

In order to solve various problems existing in infrared small target detection and improve detection efficiency, this paper focuses on improving relevant algorithms and proposes an infrared small target detection algorithm based on improved YOLOv7. The contributions are as follows:

- Design a three-scale lightweight network architecture, ISTD-YOLO, to improve the detection ability of small targets.
- The introduction of parameter-free attention mechanism SimAM improves the learning effect of the model without adding parameters to the original network structure.
- The Neck network is readjust, and a lightweight neck network named LTSN is designed to reduce the computational cost of the model, reduce the weight of the model and speed up the inference time.
- Normalized Wasserstein Distance (NWD) metric is used to optimize the standard IoU metric to improve the detection accuracy of small targets.

## 2    Related Work

Infrared small target detection has been a challenging research topic due to the presence of complex backgrounds, low signal-to-noise ratios, and the small size and weak brightness of targets. Over the years, various methods have been proposed to improve detection performance. These approaches can be broadly categorized into traditional model-driven methods and deep learning-based methods.

### 2.1   Traditional Model-Driven Methods

Early infrared small target detection methods primarily relied on model-driven approaches, which utilized handcrafted features and statistical models to distinguish targets from the background. Common techniques include: filtering-based method, local contrast-based method, sparse representation and low-rank approximation. Lee et al. [11], aiming at the problem of small target detection in infrared search and tracking system, proposed a method based on enhanced infrared intensity map and density clustering. By combining infrared intensity and standard deviation, the speed and accuracy of small target detection are improved, and the shape and centroid of the target are identified by density clustering.

### 2.2   Deep Learning-Based Methods

With the advancement of deep learning, convolutional neural networks (CNNs) and transformer-based models have demonstrated remarkable improvements in infrared small target detection. Some key developments include: CNN based detectors, lightweight network design, improved loss functions and location metrics. Gang et al. [12] proposed an improved infrared target detection SSD method based on convolutional neural networks, and the detection accuracy and effect were significantly improved by using the self-built infrared aircraft data set. Ciocarlan et al. [13] proposed the method of integrating a contrario decision criterion into the YOLO network, which could improve the accuracy of small target detection, reduce the interference of background noise, and significantly improve the performance of YOLO in the environment with few samples and limited resources.

### 2.3   Summary and Challenges

While deep learning-based approaches have significantly improved infrared small target detection, several challenges remain. Existing methods often struggle with computational efficiency, false positive suppression, and maintaining high detection accuracy in complex environments. To address these issues, this paper proposes ISTD-YOLO, a multi-scale lightweight detection algorithm based on improved YOLOv7. By integrating a lightweight network structure, an enhanced attention mechanism, and the NWD metric, ISTD-YOLO aims to achieve high-accuracy detection with reduced computational cost.

## 3   Methodology

### 3.1   Network Reconfiguration

The traditional YOLOv7 network is difficult to deal with small and weak signals and complex backgrounds in infrared images, resulting in redundant calculation and insufficient feature extraction. To address this issue, this study uses a lightweight network architecture to improve small-object feature extraction and reduce model parameters and computation, making it suitable for resource-constrained platforms.

**Lightweight Backbone Network.** Since the background dominates the infrared image and the proportion of small targets is small, the proposed model focuses more on shallow detail fea-tures and avoids overly complex feature extraction structures. Based on this, we can try to reduce the number of network convolutions and highlight the underly-ing feature maps, so as to achieve the purpose of improving the effect of model detection. The ISTD-YOLO model improves the detection efficiency by adjusting the structure of the backbone extraction network CSPDarkNet. Table 1 shows the reconstructed CSPDarkNet structure.

The advantages of reconstructing the CSPDarkNet structure are as follows:

1) Optimize multi-scale feature fusion: improve the receptive field of shallow small target features, make the model easier to use, and lightweight the original structure, which is more suitable for low-performance platform deployment..

2) Reduced network complexity: while maintaining high accuracy, the computational requirements and model size are greatly reduced, and the training and inference time is shortened.

3) Parameters are greatly reduced: the number of backbone network parameters is reduced from 13371808 to 6023584, which is only 45% of the original model.

Table 1. Reconstructed CSPDarkNet structure.

| Module | Parameters | Channel | Kernel Size | Output |
|--------|-----------|---------|-------------|--------|
| CBS    | 928       | 32      | (3,3)       | 640×640 |
| CBS    | 18560     | 64      | (3,3)       | 320×320 |
| CBS    | 36992     | 64      | (3,3)       | 320×320 |
| CBS    | 73984     | 128     | (3,3)       | 160×160 |
| ELAN   | 230656    | 256     |             | 160×160 |
| MP-1   | 213760    | 256     |             | 80×80 |
| ELAN   | 920064    | 512     |             | 80×80 |
| MP-1   | 853504    | 512     |             | 40×40 |
| ELAN   | 3675136   | 1024    |             | 40×40 |

**Adjust the Output Size of the Feature Map.** The prediction end of the original YOLOv7 model outputs three feature maps of different sizes (20×20, 40×40, 80×80 pixel) to detect large, medium, and small targets. In order to adapt to infrared small target detection, ISTD-YOLO improves YOLOv7 by deleting 20×20 pixel large receptive field feature map and adding 160×160 pixel small receptive field feature map.

ISTD-YOLO: A Multi-Scale Lightweight High-Performance Infrared Small Target Detection Algorithm    5

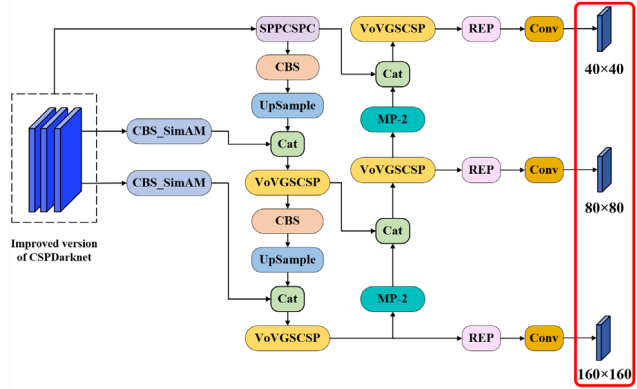

**Fig. 1.** Feature fusion network structure diagram after network reconstruction. In the figure, adjust the size of the feature map output from the prediction end of YOLOv7 model to 40 × 40, 80 × 80, 160 × 160 pixels.

The feature fusion network of YOLOv7 original model after network reconfiguration (as shown in Fig. 1), the final feature map size is 1/16, 1/8 and 1/4 of the input image respectively. This design increases the weight of shallow features to reduce the information loss of infrared dim targets caused by excessive down sampling.

Combined with the experimental verification in the fourth part, compared with the original YOLOv7, the number of parameters of the reconstructed network model is reduced by about 55%, the calculation amount is significantly reduced, and the detection accuracy is maintained at the same time, which fully explains the advantages of the lightweight network architecture design in this paper.

### 3.2   Parameter-Free Attention Mechanism

In the infrared small target detection, the small dim target occupies very few pixels in the infrared image. The attention mechanism can help the network quickly filter out the feature information of the small dim target from the complex infrared image, enhance the correlation of the local context information, and thus extract the spatial position of the target more accurately. At present, the existing attention mechanisms mainly generate one-dimensional or two-dimensional weights from feature X, and then generalize the generated weights to channel or spatial attention. However, SimAM[14] directly estimates 3D weights, and the structural principle is shown in Fig. 2.

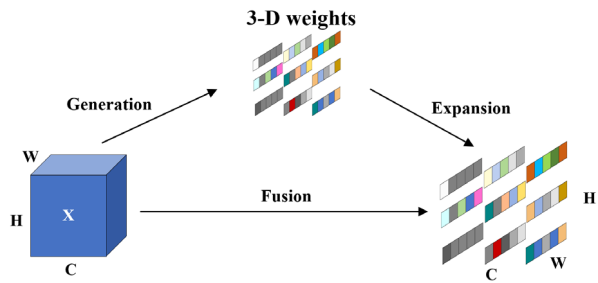

**Fig. 2.** SimAM Structural schematic diagram. The same color represents a single scalar used for each channel, spatial location or each point on the feature.

In computer vision, Common Attention modules such as SE (Squeeze-and-Excitation) [15], CBAM (Convolutional Block Attention Module) [16], and GC (Global Context) [17] focus on the channel domain or spatial domain. It corresponds to the feature-based or space-based attention mechanism of the human brain. However, SimAM simulates the human brain attention mechanism to generate realistic 3D weights, and assigns a unique weight to each neuron. First, we define an energy function for each neuron:

$$e_t^* = \frac{4(\hat{\sigma}^2 + \lambda)}{(t - \hat{\mu})^2 + 2\hat{\sigma}^2 + 2\lambda} \quad (1)$$

Where: $\hat{\mu} = \frac{1}{M}\sum_{i=1}^{M} X_i$, $\hat{\sigma}^2 = \frac{1}{M}\sum_{i=1}^{M}(x_i - \hat{\mu})^2$, $M = H \times W$ is the number of neurons on this channel, $i$ is the index in the spatial dimension, $t$ and $x_i$ denote the target neuron and the other neurons in the input feature single channel, respectively. According to Equation (1), the lower the energy $e_t^*$, the more different the neuron t is from its surrounding neurons and the more important it is for visual processing. The importance of each neuron can be obtained by taking the inverse of $e_t^*$. The features are then refined using the scaling operator, and the entire refinement phase is as follows:

$$\tilde{X} = sigmoid\left(\frac{1}{E}\right) \odot X \quad (2)$$

$E$ is the grouping of all $e_t^*$ in channel and spatial dimensions, and sigmoid is present to limit too large values in $E$. *sigmoid* is a single function that does not affect the relative importance of each neuron.

Compared with the existing channel and spatial attention modules, SimAM can infer the 3D attention weights of the feature maps in the layer without adding parameters to the original network structure. In order to improve the accuracy and robustness of dim and small target detection, parameter-free attention mechanism SimAM is introduced after the first two CBS of the neck network, respectively. The Grad-CAM method is used to generate the attention heat map on the test result map, as shown in Fig. 3.

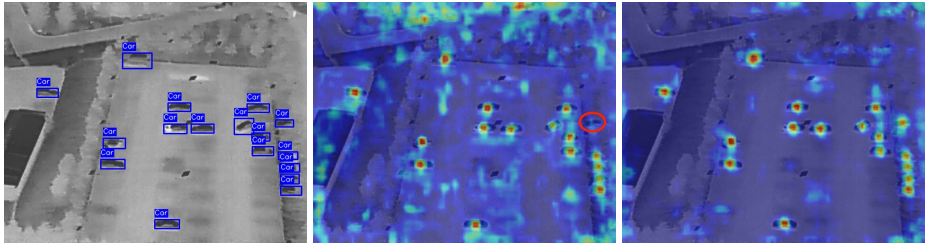

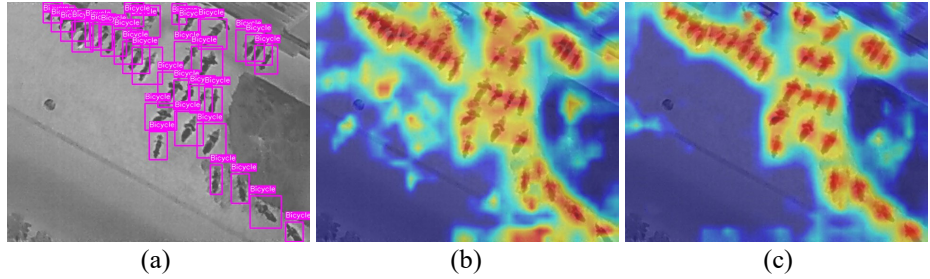

(a)    (b)    (c)

**Fig. 3.** Comparison of the distribution of attention regions. The picture in column (a) is the Ground Truth, the picture in column (b) is the thermal map of target detection of YOLOv7 model, and the picture in column (c) is the thermal map of target detection of YOLOv7+SimAM model.

The analysis of the image shows that the target detection heat map of the original YOLOv7 model has missed detection (which has been marked with the red circle). After the introduction of SimAM, the model's ability to capture the target feature information is significantly improved, and the overall attention of the network is focused more on the target area, which has superior detection performance compared with the original network.

### 3.3  Lightweight Neck Network

In order to complete the real-time detection task of infrared small target and achieve satisfactory detection accuracy, this paper re-adjusts the Neck network of the model based on the reconstructed feature extraction network, and designs a Lightweight Three Scale Neck (LTSN) network that is more suitable for small target detection. On the basis of scale transformation, LTSN replaces the ELAN-W module in the structure with the lighter single aggregation module VoVGSCSP[18]. It uses a new convolution operator GSConv, which reduces the computational cost of the model and achieves significant accuracy gain without additional operations.

**Lightweight Convolution.** In order to make the output of Depthwise Separable Convolution (DSC) as close as possible to the Standard Convolution (SC), so that the lightweight model built by a large number of DSC can achieve higher accuracy, and let the huge model complete the task of real-time detection, Li et al. proposed a new lightweight convolution technology, which is a hybrid convolution of SC, DSC and shuffle. Named GSConv, its structural composition is shown in Fig. 4. Aiming at the special requirements of infrared small target detection tasks, this paper considers the introduction of lightweight convolution GSConv. The experimental results show that the model using GSConv method reduces the negative impact of DSC defects on detection, reduces the weight of the model, reduces the computational cost and complexity, and improves the detection accuracy.

8    S. Zhang et al.

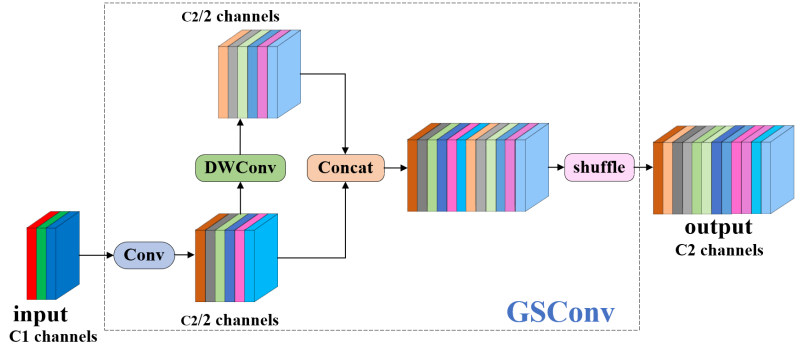

**Fig. 4.** GSConv structure diagram.

**Single Aggregation Module.** Li et al. designed the one-shot aggregation based cross-stage Partial Network (GSCSP) module and proposed three design schemes for VoV-GSCSP (Fig. 5 (a)(b)(c)). Among them, (a) has the simplest structure and the fastest inference time, while (b) and (c) have higher feature reuse rates. Since the simple structure is more friendly to hardware, this paper chooses the more cost-effective scheme (a). In the VoV-GSCSP module, the input feature map is divided into two branches.

In order to reduce the inference time and improve the inference speed, this paper considers using a single-shot aggregation module VoV-GSCSP to replace the ELAN-W module in the neck of the model to improve the inference speed of the algorithm. VoV-GSCSP can reduce the amount of computation and complexity of the network structure, while maintaining sufficient accuracy.

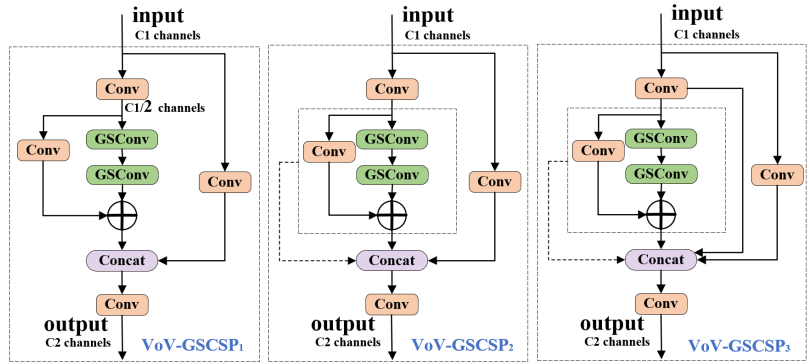

**Fig. 5.** VoV-GSCSP structure diagram.

### 3.4   Normalized Gaussian Wasserstein Distance

In object detection, YOLOv7 uses Intersection over Union (IoU) to measure the similarity between the real box and the predicted box. However, IoU is sensitive to the position deviation of small targets, which will also reduce the performance in the anchor based detector. In order to solve this problem, this paper introduces Normalized Wasserstein Distance (NWD) as a new measurement method [19]. In order to better distinguish the weights of different pixels in the bounding box, the bounding box can be modeled as a two-dimensional Gaussian distribution, and the Wasserstein distance is used to measure the distribution similarity, so as to optimize the standard IoU. The horizontal bounding box is built as a two-dimensional Gaussian distribution $N(\mu,\Sigma)$, where $\mu$ and $\Sigma$ denote the mean vector and covariance matrix, respectively.

For two 2D Gaussian distributions $\mu_1=N(m_1, \Sigma_1)$ and $\mu_2=N(m_2, \Sigma_2)$, the second-order Wasserstein distance between them is defined as follows.

$$W_2^2(\mu_1,\mu_2) = \|m_1 - m_2\|_2^2 + Tr\left(\Sigma_1 + \Sigma_2 - 2\left(\Sigma_2^{1/2}\Sigma_1\Sigma_2^{1/2}\right)^{1/2}\right) \quad (3)$$

The formula can be simplified as follows.

$$W_2^2(N_a,N_b) = \left\|\left(\left[cx_a, cy_a, \frac{w_a}{2}, \frac{h_a}{2}\right]^T, \left[cx_b, cy_b, \frac{w_b}{2}, \frac{h_b}{2}\right]^T\right)\right\|_2^2 \quad (4)$$

Where, $(c_x,c_y)$ is the center coordinate of the bounding box; w and h denote the width and height.

However, $W_2^2(N_a,N_b)$ is a distance metric, which does not directly measure the similarity between two bounding boxes, so we can normalize it in exponential form to obtain a new metric NWD, which is the normalized Wasserstein distance:

$$NWD(N_a, N_b) = exp\left(-\frac{\sqrt{W_2^2(N_a,N_b)}}{C}\right) \quad (5)$$

Where, C is a constant closely related to the data set. The NWD metric is designed as a loss function:

$$L_{NWD} = 1 - NWD(N_p, N_g) \quad (6)$$

Where, $N_p$ is the Gaussian distribution model for predicting the bounding box $P$; $N_g$ is the Gaussian distribution model of the true bounding box $G$, and the NWD-based metric loss can provide gradients even in both cases $|P \cap G|=0$ and $|P \cap G|=P$ or $G$.

The loss function of YOLOv7 includes classification loss, localization loss and regression loss, and CIoU is used as the bounding box regression loss lbox, which is calculated as follows.

$$l_{box} += (1.0 - iou)mean(\ ) \quad (7)$$

Where, iou represents the overlap rate between the predicted bounding box and the true bounding box, and mean() represents the mean function.

In order to improve the detection performance of small objects, this paper introduces the Normalized Gaussian Wasserstein distance (NWD) into the regression loss, and the improved formula is:

$$l_{box} += (1.0 - iou\_ratio)(1.0 - nwd)mean(\ ) + iou\_ratio(1.0 - iou)mean(\ ) \quad (8)$$

Where, iou_ratio represents the proportion of IoU metrics to balance the weights of nwd and original iou. nwd represents the similarity loss between the predicted bounding box measured by the normalized Gaussian Wasserstein distance and the true bounding box.

The benefit of using the Wasserstein distance is that it measures the similarity of distributions even when there is no overlap between the predicted bounding box P and the true bounding box G (that is, $|P \cap G|=0$) or the overlap is rarely negligible. In addition, NWD is not sensitive to objects of different scales, so it is more suitable for the detection of small objects.

### 3.5   The ISTD-YOLO Structure

In order to accurately detect small targets, the network structure of YOLOv7 is reconstructed and lightweight. Firstly, the CSPDarkNet structure of YOLOv7 backbone network was adjusted, and the feature map output of 20×20 pixel scale was removed at the preresetting end of the model, and the feature map output of 160×160 pixel size was added to improve the weight of small receptive field and strengthen the attention to shallow detail features. Secondly, after the first two CBS modules of the neck network, the parameter-free attention mechanism SimAM is introduced to form the CBS_SimAM module, which enhances the learning ability of the network without adding parameters to the original network structure. Finally, a lightweight neck network LTSN is designed, and a new convolution technique GSConv is introduced, which replaces ELAN-W with VoV-GSCSP to reduce the amount of calculation and parameters of the model, so as to achieve the purpose of lightweight neck network. The ISTD-YOLO network structure is shown in Fig. 6.

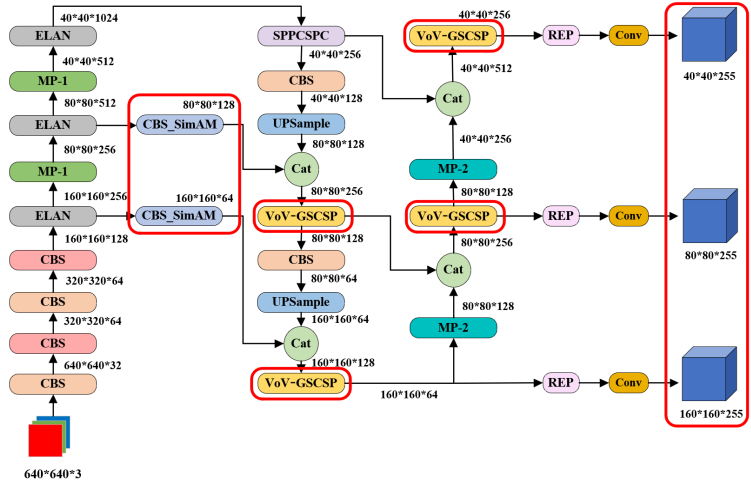

**Fig. 6.** Detail of the ISTD-YOLO algorithm structure. Firstly, the YOLOv7 network is reconstructed, that is, the CSPDarkNet structure of the backbone network is adjusted and the output size of the feature graph of the prediction terminal is adjusted. Secondly, after the first two CBS modules of the neck network, SimAM is introduced to form CBS_SimAM module. Finally, the ELAN-W module in the original network structure is replaced by VoV-GSCSP.

While the lightweight design significantly improves computational efficiency, it may sometimes compromise feature richness under extremely complex conditions. The SimAM attention mechanism, by assigning unique weights to each neuron without extra parameters, offers computational efficiency but its adaptability across different scales requires further investigation. The NWD metric effectively solves the sensitivity problem of IoU in small target detection. However, its performance in the case of heavy target overlap still needs to be improved.

## 4 Experiment

### 4.1 Experiment Environment and Dataset Selection

**Experiment Environment.** The hardware platform and environmental parameters used by the experimental training model are shown in Table 2.

Table 2. Hardware platform and environment parameters.

| Parameters | Configuration&Setup |
| --- | --- |
| System | Windows10 |
| CPU | Intel(R) Xeon(R) Silver 4310 |
| GPU | NVIDIA GeForce RTX 3090, 24G |
| PyTorch | Pytorch1.12.1, CUDA11.6.0 |
| Python | Python3.7 |
| Input image size | 640×640 |
| Batch Size | 16 |
| Epochs | 300 |

**Dataset Selection.** HIT-UAV[20] and IRSTD-1k[21] were selected as data sets for the experiment.

HIT-UAV is the world's first publicly released high-altitude human-vehicle infrared imaging dataset based on Unmanned Aerial vehicle (UAV), which contains 2898 infrared thermal images extracted from hundreds of videos taken by UAV in different scenes (such as schools, parking lots, roads, playgrounds, etc.). Each image in the dataset is 640×512 pixels in size. In this paper, 70% of the images are selected as the training set, 20% as the verification set, and 10% as the test set.

The IRSTD-1k dataset consists of 1,000 real images with various object shapes, different object sizes, and clutter-rich backgrounds, with image dimensions of 512×512 pixels and backgrounds with accurate pixel-level annotations. IRSTD-1k contains different types of small targets such as drones, creatures, boats, and vehicles, and this dataset covers many different scenes with backgrounds including ocean, river, field, mountain, city, and cloud with severe clutter and noise. In this paper, 70%, 15% and 15% of the images are randomly selected as the training set, validation set and test set of the experiment, respectively.

### 4.2   Evaluation Index

Precision(P), Recall(R), and mean Average Precision at IoU threshold 0.5 (mAP0.5) are used as evaluation metrics in this experiment. In addition, we perform a comprehensive evaluation of model performance in terms of number of parameters, model size, and GFLOPs.

### 4.3   Ablation Experiment

The HIT-UAV dataset was selected in the experiment, YOLOv7 was used as the benchmark algorithm, and P, R, mAP0.5, Model Size, Parameters, and GFLOPs were used as evaluation indicators. The experimental results are shown in Table 3.

In complex background, weak target and high noise scenes, ISTD-YOLO shows excellent robustness. Experimental results show that ISTD-YOLO has significantly higher precision and recall than YOLOv7 under these extreme conditions, which proves the positive impact of using lightweight network architecture and introducing SimAM attention mechanism on small object detection. At the same time, after network reconstruction and lightweight processing, the model size is greatly reduced, the number of parameters and GFLOPs are significantly reduced, and the recall rate is also improved after using LTSN or introducing NWD. Therefore, the experimental indicators of ISTD-YOLO are significantly improved compared with the original YOLOv7 algorithm, which verifies the superiority of ISTD-YOLO.

Table 3. HIT-UAV dataset ablation experiments.

| Model | Reconfiguration | SimAM | LTSN | NWD | P/% | R/% | mAP0.5/% | Model Size/MB | Parameters/M | GFLOPs |
|---|---|---|---|---|---|---|---|---|---|---|
| YOLOv7s | | | | | 72.91 | 66.53 | 69.32 | 71.3 | 37.2 | 105.1 |
| A | √ | | | | 76.15 | 68.85 | 71.92 | 23.8 | 12.3 | 99.1 |
| B | | √ | | | 75.36 | 68.04 | 72.87 | 71.3 | 37.2 | 105.1 |
| C | | | √ | | 76.34 | 70.52 | 72.21 | 61.2 | 31.9 | 91.9 |
| D | | | | √ | 76.03 | 70.02 | 72.09 | 71.3 | 37.2 | 105.1 |
| E | √ | √ | | | 76.38 | 71.43 | 73.65 | 23.9 | 12.3 | 99.1 |
| F | √ | | √ | | 77.19 | 72.02 | 74.36 | 21.8 | 11.2 | 86.6 |
| G | √ | | | √ | 76.29 | 71.58 | 73.97 | 23.9 | 12.3 | 99.1 |
| H | | √ | √ | | 75.86 | 70.92 | 74.08 | 61.6 | 32.1 | 92.8 |
| I | | √ | | √ | 76.89 | 72.61 | 74.36 | 71.3 | 37.2 | 105.1 |
| J | | | √ | √ | 77.37 | 71.57 | 73.82 | 61.6 | 32.1 | 92.8 |
| K | √ | √ | √ | | 80.96 | 72.90 | 75.03 | 21.8 | 11.2 | 86.6 |
| L | √ | | √ | √ | 83.65 | 73.01 | 75.61 | 21.8 | 11.2 | 86.6 |
| M | √ | √ | | √ | 82.39 | 72.64 | 74.91 | 23.9 | 12.3 | 99.1 |
| ISTD-YOLO | √ | √ | √ | √ | 88.77 | 74.20 | 77.84 | 21.8 | 11.2 | 86.6 |

ISTD-YOLO: A Multi-Scale Lightweight High-Performance Infrared Small Target Detection Algorithm 13

## 4.4 Comparative Experiment

**Comparison of Different Backbone Networks.** ISTD-YOLO reconstructs the original network model of YOLOv7, and achieves the effect of lightweight network model: the structure of the backbone feature extraction network CSPDarkNet is adjusted and the output size of the feature map is adjusted to 40×40, 80×80, and 160×160 pixels. In this experiment, in order to study the significant effect of network reconstruction for infrared small target detection, we will use lightweight networks ShuffleNetV2[22] and MobileNetV3[23] to replace the CSPDarkNet structure in the backbone network, and the feature map output size remains the above output. MobileNetV3 is available in Large and Small versions. To better capture the characteristic information of tiny targets, we choose to use the MobileNetV3-Large version.

The experimental results are shown in Table 4. According to the comparative experimental results, it is easy to see that although the computational complexity of the model reconstructed by the network in this paper is slightly higher, the model parameters and model volume are significantly smaller than the other two models, and the detection accuracy also maintains a good effect. It is verified that the model reconstructed by the network in this paper realizes a good lightweight design and is more suitable for the detection of small infrared targets.

**Table 4.** Comparison of different backbone networks on the HIT-UAV dataset.

| YOLOv7s | Model Size /MB | mAP0.5 /% | Parameters /M | GFLOPs |
|---|---|---|---|---|
| + ShuffleNetV2 | 44.1 | 69.8 | 25.8 | 87.5 |
| + MobileNetV3-L | 51.4 | 71.7 | 25.5 | 41.1 |
| + Reconfiguration | 23.8 | 71.9 | 12.3 | 99.1 |

**Comparison of Different Algorithms.** In order to verify the effectiveness of ISTD-YOLO for infrared small target detection, it was compared with various popular YOLO series algorithms for target detection on the HIT-UAV dataset. As shown in Table 5, the comparison experiments show that the size of ISTD-YOLO model is 21.8 MB, the mAP0.5 reaches 77.84%, which is the best performance, and the inference time is reduced by 13.5%. In general, ISTD-YOLO has the advantages of lightweight and high accuracy, which is suitable for efficient deployment on platforms with limited hardware resources.

**Table 5.** Comparison of algorithms for the HIT-UAV dataset.

| Model | Model Size /MB | mAP0.5/% | Inference time /ms |
|---|---|---|---|
| YOLOv5s | 13.7 | 61.35 | 5.8 |
| YOLOv6s | 38.7 | 60.03 | 10.9 |
| YOLOv7s | 71.3 | 69.32 | 13.3 |
| YOLOv7-tiny | 11.7 | 59.84 | 5.2 |
| YOLOv8s | 21.4 | 72.45 | 5.7 |
| YOLOv8n | 5.96 | 70.06 | 4.9 |
| ISTD-YOLO | 21.8 | 77.84 | 11.5 |

### 4.5  Multi-Class Object Detection Capability

To evaluate the detection performance of our model, we can visualize the results using a confusion matrix, where each row represents the actual class, each column represents the predicted class, the diagonal values are the proportion of correct predictions, and the values in the last row and last column represent the false positive rate and miss rate for each class, respectively. As shown in Fig. 7, the accuracy of the improved algorithm is higher than that of the original algorithm, and the false detection rate and missed detection rate are significantly reduced. For the OtherVehicle and DontCare categories that are difficult to detect in the data set, the correct prediction proportion of the improved model is increased by 6% and 20% respectively. In general, ISTD-YOLO has good multi-scale detection performance in multi-class object detection.

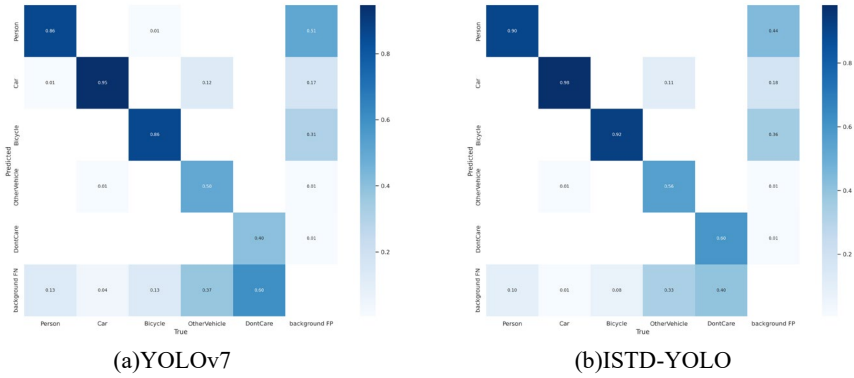

(a)YOLOv7                                         (b)ISTD-YOLO

**Fig. 7.** Confusion Matrix. Figure (a) shows the results of the original YOLOv7 model and Figure (b) shows the results of the improved model.

### 4.6  Experimental Results and Analysis

In order to verify the detection effect of ISTD-YOLO algorithm in the actual application environment, this paper will randomly select from the HIT-UAV test set images for detection. The comparison of the final detection effect is shown in Fig. 8. It can be seen that YOLOv7 has the problems of false detection and missed detection, and the detection accuracy of ISTD-YOLO is higher than that of the original model. It is proved that the improved model performs well in infrared images with complex background and weak target brightness, has outstanding background suppression ability, effectively improves the processing ability of complex samples, and reduces the false detection rate and missed detection rate.

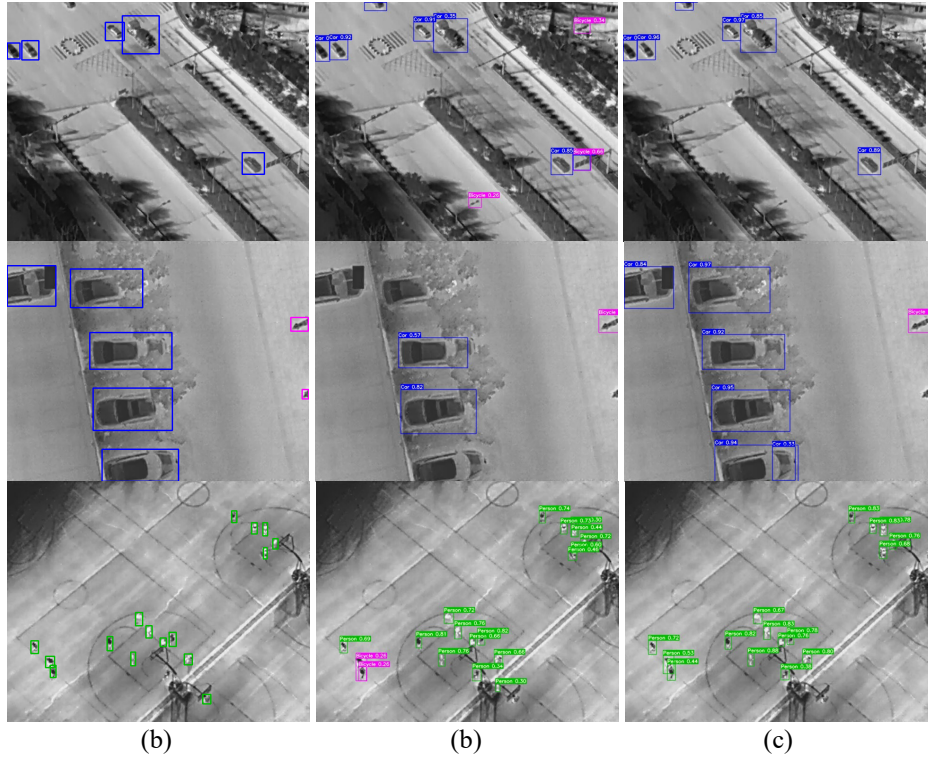

(b)　　　　　　　(b)　　　　　　　(c)

**Fig. 8.** Comparison of detection effect on HIT-UAV dataset. (a) Column label visual image, where green box is labeled Person, blue box is labeled Car, pink box is labeled Bicycle; (b) column shows the detection effect of YOLOv7s; (c) Column shows the effect of ISTD-YOLO detection.

### 4.7　Model Generalization Verification

**Table 6.** Comparison of detection results on the IRSTD-1k dataset.

| Model | P% | R% | mAP0.5/% | Parameters /M | GFLOPs |
|---|---|---|---|---|---|
| YOLOv3s | 77.2 | 73.4 | 78.2 | 103.7 | 283.0 |
| YOLOv5s | 81.5 | 73.0 | 79.3 | 2.5 | 7.2 |
| YOLOv7s | 78.6 | 65.9 | 68.1 | 37.2 | 105.1 |
| YOLOv8s | 81.8 | 76.0 | 79.2 | 3.0 | 8.2 |
| ISTD-YOLO | 84.0 | 75.5 | 79.5 | 11.2 | 86.6 |

In order to verify the significant test effect and good generalization of ISTD-YOLO model on other public infrared small target datasets, this paper chooses to conduct comparative tests on the public dataset IRSTD-1k. The experimental results are shown in Table 6. Compared with the original YOLOv7 model, the Precision, Recall and mAP50 of the improved model have increased by 5.4%, 9.6% and 11.4% respectively. At the same time, compared with other algorithms, the mAP50 improvement effect of the proposed algorithm is obvious, which verifies the good robustness and generalization of the ISTD-YOLO model.

Fig. 9 shows the comparison of the detection effect of the algorithm before and after the improvement. It is easy to see that the baseline algorithm has the problem of false detection, and the detection accuracy of the improved model is higher than that of the original model. The ISTD-YOLO model in this paper shows excellent performance in scenes with complex backgrounds and weak targets, which verifies the effectiveness and generalization of the model.

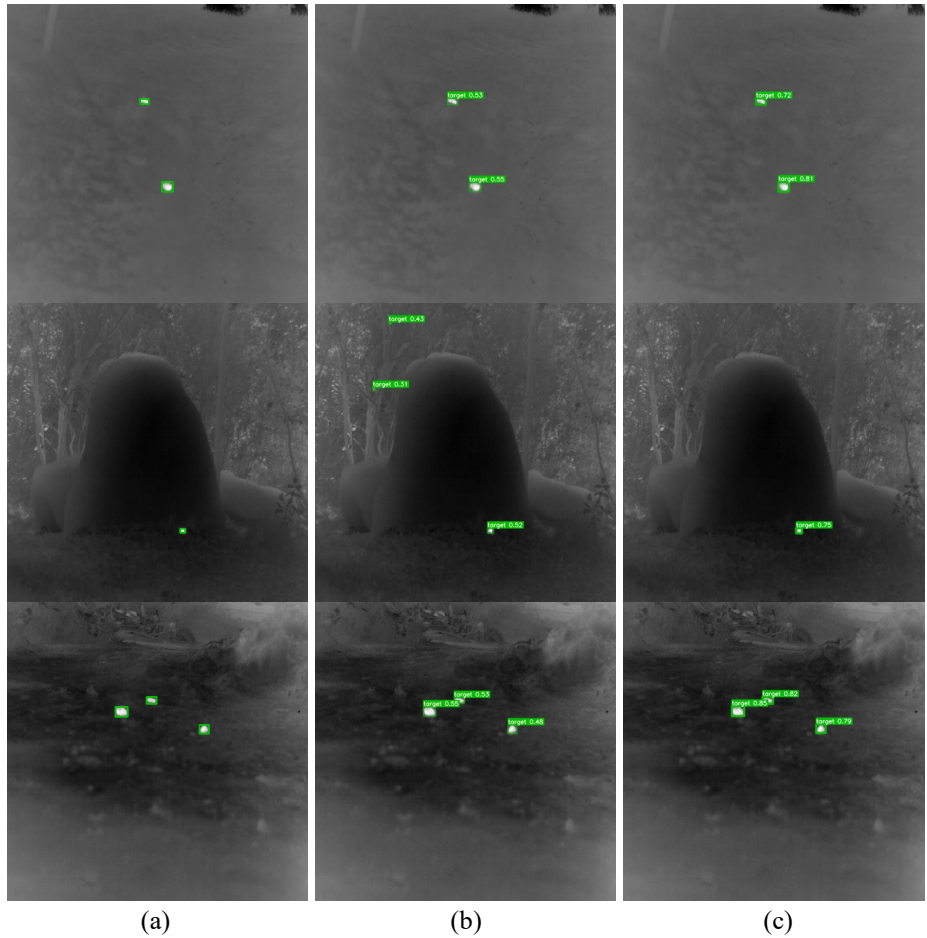

(a)                (b)                (c)

**Fig. 9.** Comparison of detection effects on IRSTD-1k dataset. Column (a) shows the labeled visualization image, column (b) shows the detection results of YOLOv7, and column (c) shows the detection results of ISTD-YOLO.

## 5  Conclusions

Infrared small target detection is a hot and difficult problem in the field of target detection. In this paper, an ISTD-YOLO algorithm based on improved YOLOv7 is proposed. The innovations include: Firstly, the network reconstruction of YOLOv7 model is performed to improve the utilization of shallow features and realize multi-scale lightweight detection. Secondly, the parameter-free attention mechanism SimAM is introduced to make the model focus on the effective target features. Thirdly, a lightweight neck network LTSN is designed by using VoV-GSCSP instead of ELAN-W and combining with GSConv convolution technology, which reduces the weight and computational complexity of the model. Finally, a new object localization metric, NWD, is introduced to optimize the IoU metric to improve the performance of small object detection. The experimental results show that ISTD-YOLO can achieve high-quality detection of infrared small targets, and has the advantages of real-time, low computational cost and high detection accuracy in practical applications. Given the rapid development of object detection technology, future research may focus on further optimizing the balance between lightweight design and new metrics, explore the robustness improvement in more complex scenarios, and the possibility of extending this method to other object detection tasks.